\documentclass{article}
\usepackage{spconf,amsmath,graphicx}

\usepackage{amsfonts,amssymb,url,gensymb}
\usepackage{tabularx,multirow,arydshln,booktabs,hyperref}
\usepackage{xcolor}
\usepackage{fancyhdr}

\newif\ifblind
\blindfalse 

\fancypagestyle{firstpage}{%
  \fancyhf{}
  \fancyhead[L]{\small © 2025 IEEE. Personal use of this material is permitted. Permission from IEEE must be obtained for all other uses, in any current or future media, including reprinting/republishing this material for advertising or promotional purposes, creating new collective works, for resale or redistribution to servers or lists, or reuse of any copyrighted component of this work in other works.}
  
  \setlength{\headheight}{25pt}
  \setlength{\topmargin}{-0.3in}
}

\title{User-in-the-Loop View Sampling with Error Peaking Visualization}
\ifblind
    \name{Author(s)}
    \address{Affiliation(s)}
\else
    \name{Ayaka Yasunaga$^1$, Hideo Saito$^1$, and Shohei Mori$^{2,1}$}
    \address{$^1$Keio University~~$^2$University of Stuttgart}
\fi

\begin{document}

\thispagestyle{firstpage}

\maketitle
\begin{abstract}
Augmented reality (AR) provides ways to visualize missing view samples for novel view synthesis. Existing approaches present 3D annotations for new view samples and task users with taking images by aligning the AR display. This data collection task is known to be mentally demanding and limits capture areas to pre-defined small areas due to the ideal but restrictive underlying sampling theory. To free users from 3D annotations and limited scene exploration, we propose using locally reconstructed light fields and visualizing errors to be removed by inserting new views.
Our results show that the error-peaking visualization is less invasive, reduces disappointment in final results, and is satisfactory with fewer view samples in our mobile view synthesis system. We also show that our approach can contribute to recent radiance field reconstruction for larger scenes, such as 3D Gaussian splatting.
\end{abstract}
\begin{keywords}
Augmented reality, multi-plane images, user-in-the-loop, error peaking, view synthesis.
\end{keywords}
\section{Introduction}
\label{sec:intro}

Novel view synthesis can render unseen views from given multi-viewpoint images. Recent advances in machine learning provide a way to encode scene appearance into a trained neural network \cite{mildenhall2020nerf,wang2021ibrnet,sitzmann2021lfns} or to optimize scene proxies \cite{fridovich2022plenoxels,tucker2020single,kerbl20233dgs}. This process takes minutes to hours, and missing contributing view samples will lead to another photography session and optimization. Plenoptic sampling theory provides a guideline for taking the minimum number of images for aliasing-free view synthesis \cite{chai2000plenoptic}. However, it only supports view sampling in 2D planes or light fields (LF) \cite{LevoyHanrahan1996LightField}. A practical view sampling session will require thousands of images even with an approximation by multi-plane images (MPI) \cite{mildenhall2019local}.

Augmented reality (AR) plays an important role in visualizing missing view samples at pre-calculated locations in 3D space. Existing work proposed a data collection task, which measures the minimum scene depth for plenoptic sampling theory and places 3D annotations \cite{mildenhall2019local}. The user is asked to move a tracked smartphone to the annotations to trigger shutters. Similar approaches can apply to spherical visualization \cite{mohr2020mixed,davis2012unstructured}. The issues of AR visual guidance are
(i) the visualizations hiding the subject to be photographed, hindering the entertaining nature of photography,
(ii) the alignment task that needs to be precise and is mentally demanding \cite{ishikawa2023multi},
(iii) the globally applied minimum view intervals that may not make sense at the moment, and
(iv) fixed visualization limiting users' ability to explore beyond predetermined areas.

\begin{figure}[!t]
    \centering
    \includegraphics[width=\columnwidth]{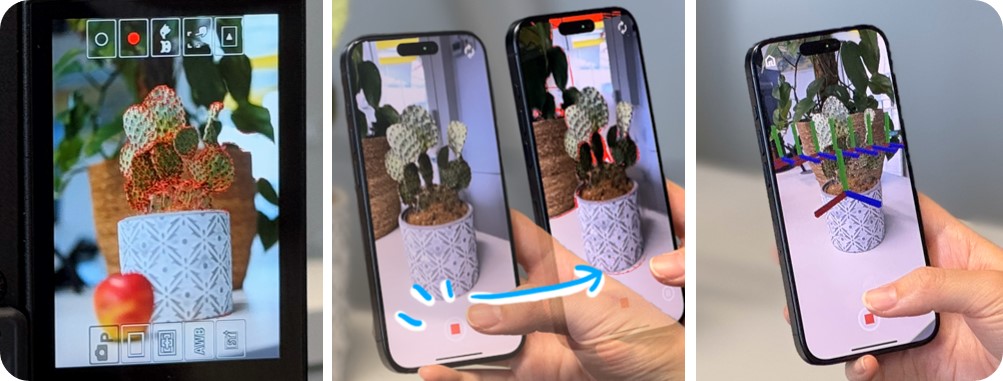}
    \caption{Focus peaking, our error peaking, and conventional AR visualizations.
    Inspired by focus peaking, our approach visualizes errors in the underlying view synthesis, encouraging the insertion of new views to diminish noticeable errors. Compared to the conventional AR visualization, our method is less invasive and addresses existing issues.
    }
    \label{fig:error-peaking}
\end{figure}

To resolve these issues, we propose error-peaking visualization using local light field reconstruction or MPI \cite{mildenhall2019local,han2022adaMPI}. Similar to the focus peaking of digital cameras that visualize focused areas to let the photographer find the best in-focus areas, our error-peaking visualization highlights erroneous pixels that are diminished by inserting more view samples (Fig.~\ref{fig:error-peaking}). Overall, the user completes a photographing session when the view synthesis appears with no artifacts. This visualization technique (i) disturbs artifacts only, (ii) requires no alignment task, (iii) enables the user to analyze local reconstruction, and (iv) allows free exploration in 3D space. A demonstration video is available at \href{https://sigport.org/documents/user-loop-view-sampling-error-peaking-visualization}{IEEE SigPort}.

\section{Related Work}

\subsection{Light Fields and Recording}

LF consists of 4D data of 2D locations ($uv$-plane) and a focus plane ($st$-plane) \cite{LevoyHanrahan1996LightField} and is rendered by resampling 4D data into 2D. This 4D array data can be recorded systematically with a gantry \cite{LevoyHanrahan1996LightField}, lenslet \cite{Ng05a}, or a camera array system \cite{Vaish04usingplane}. LF is an approximated plenoptic function \cite{Adelson1991plenoptic}, and thus, the number of view samples for a certain area is explained by the plenoptic sampling theory \cite{chai2000plenoptic}. More recent approaches can accept unstructured view samples assuming spatial oversampling \cite{davis2012unstructured} or more geometric information \cite{buehler2001unstructured}.

MPI is considered an approximated LF and allows sparser view sampling with soft 3D reconstruction of RGB+$\alpha$ layers \cite{mildenhall2019local,penner2017Soft3D}. MPI with $D$ layers reduces the number of view samples by $D^2$ compared to the original LF \cite{mildenhall2019local}.

\subsection{Modern Multi-Plane Images}
MPI is generated via multi-view analysis \cite{mildenhall2019local,penner2017Soft3D,szeliski1999stereo}, but modern approaches can infer MPI from a single snapshot in a few seconds using a neural network trained with enormous image datasets \cite{tucker2020single,han2022adaMPI}.
These approaches adjust layer locations to maximize the pixel densities per layer \cite{han2022adaMPI} and inpaint unobserved pixels when the rendering viewpoint diverges from the original viewpoint \cite{tucker2020single}. The modifications go beyond the support of the plenoptic sampling theory of uniform sampling.

Nevertheless, multi-view sampling is needed for broader scene coverage and to reproduce view-dependent effects. Our error-peaking visualization approach compares the MPI rendering with the current view, providing visual analysis regardless of MPI generators and sampling theory. The decision to insert a new view depends solely on the user's visual analysis.

\subsection{AR for View Sampling}
\label{sec:ar_vis}

AR can offer 3D annotations to visualize where to add view samples in the air \cite{mildenhall2019local, ishikawa2023multi}. Popular approaches include 3D axes \cite{mildenhall2019local}, 2D planes \cite{ishikawa2023multi}, and a hemisphere \cite{davis2012unstructured,mohr2020mixed} surrounding a target area. These methods (i) can partly or entirely hide the area of interest and (ii) retarget the photographing task to a data collection task. The overall experience can be mentally demanding, especially for smaller areas \cite{ishikawa2023multi}.

The locations of the visual guidance are calculated according to the plenoptic sampling theory \cite{chai2000plenoptic}. This process (iii, iv) involves entire scene scanning to guarantee the minimum scene distance during view sampling, and therefore, 3D annotations must be fixed once the session starts.

To find the next best view samples online, geometric information gain is evaluated \cite{li2025activesplat}, which ignores the view-dependent nature. The state-of-the-art online 3D Gaussian optimization integrated into simultaneous localization and mapping (SLAM) systems suppresses view-dependency and anisotropic Gaussians for speed and robustness \cite{matsuki2024gsslam, keetha2024splatam}. To gain the missing view dependency, another full optimization has to run afterward \cite{keetha2024splatam}.

We use MPI, which appears to be pixel-accurate reconstruction in a local area, enabling local visual evaluation by directly comparing the rendering with the current view. This is hard to achieve with a volumetric or 3D Gaussian reconstruction at low resolution for presenting tentative visual feedback on mobile devices, although these solutions are popular in consumer applications\footnote{For example, Scanniverse app \url{https://scaniverse.com}}.

\begin{figure}[!t]
    \centering
    \includegraphics[width=\columnwidth]{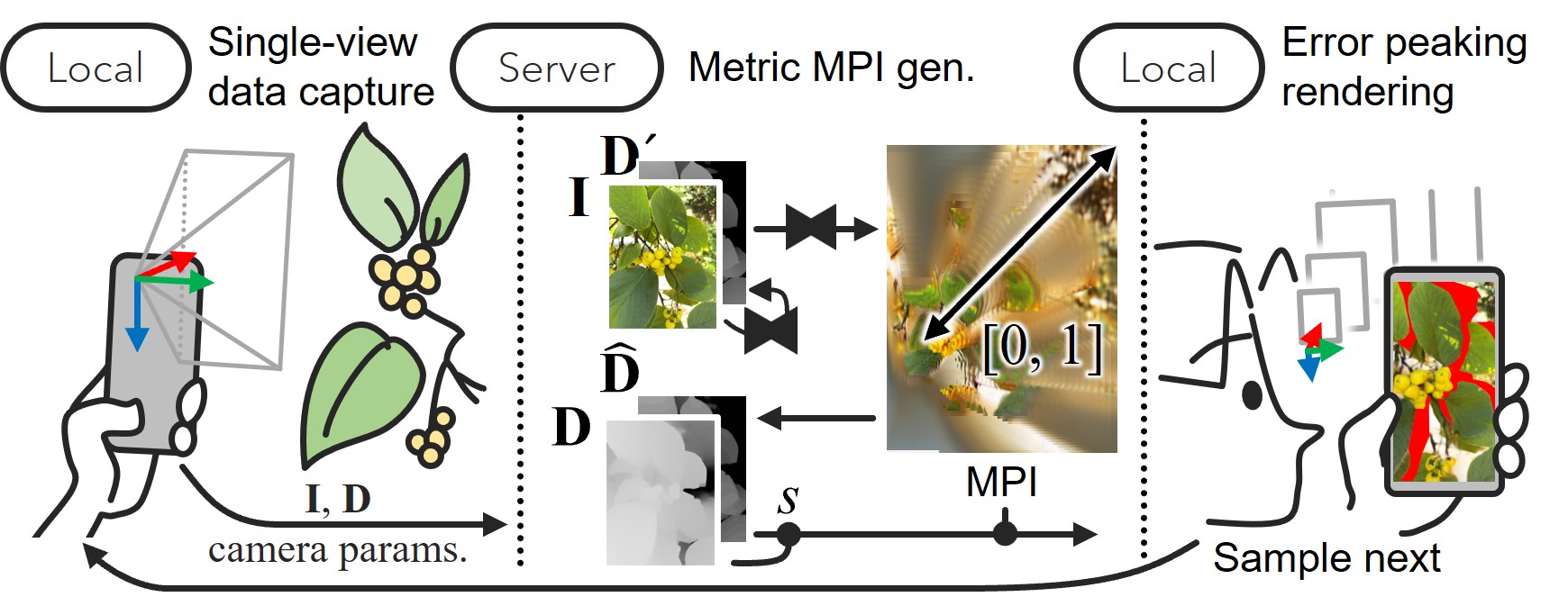}
    \caption{User-in-the-loop view sampling system overview.
    }
    \label{fig:proposed_system}
\end{figure}


\section{User-in-the-Loop LF Capture System}
Our system enables error-peaking visualization in three steps (Fig.~\ref{fig:proposed_system}): Single-view data capture with a tracked smartphone, MPI generation from a single image (Sect. \ref{sec:mpi_generation}), and per-pixel evaluation for the user's assessment (Sect. \ref{sec:vis_feedback}).

\subsection{Metric MPI Generation}\label{sec:mpi_generation}

\noindent
\textbf{Single-view data.}
Our system allows users to capture single-view data on an AR-supported mobile phone.
Single-view data consists of an RGB image $\mathbf{I}$, a metric depth map $\mathbf{D}$, and intrinsic and extrinsic camera parameters.
Given the data, the system predicts MPI from the inputs for instant visual feedback.
Assuming the AR device does not have enough computational resources to perform CNN-based MPI generation, the data is routed over a network to a server capable of more powerful computing.
In practice, the camera parameters and metric depth map are provided by Unity AR Foundation\footnote{Unity AR Foundation \url{https://unity.com/solutions/xr/ar}} using their SLAM system with an inertial measurement unit (IMU).

\noindent
\textbf{MPI from single-view data.}
Given $\mathbf{I} \in \mathbb{R}^{W \times H \times 3}$ and $\mathbf{D} \in \mathbb{R}^{W \times H \times 1}$ of size $W$ (width) by $H$ (height), we generate an MPI of $W \times H \times 4D$ using AdaMPI \cite{han2022adaMPI}, a state-of-the-art approach trained for general scenes.
MPI has $D$ layers ($D=32$ by default), and each layer consists of pixels with colors $\mathbf{c}_i$ and density $\sigma_i$ at a layer depth $(i \in [1,D])$. $i=1$ and $i=D$ are the closest and furthest layer indices, respectively.

Rendering a novel view with MPI involves volume rendering or tomography warping and compositing individual layers at a view.
Given $\mathbf{c}_i$ and $\sigma_i$ at $i_\text{th}$ MPI layer at the novel viewpoint, we calculate colors $\hat{\mathbf{c}}_i$ as follows:
\begin{equation*}
    \hat{\mathbf{c}} = \sum_{i=1}^D \left( \mathbf{c}_i \hat{\alpha}_i \prod_{j=1}^{i-1} \left( 1-\hat{\alpha}_j\right) \right),
    \hat{\alpha} = \sum_{i=1}^D \left( \alpha_i \prod_{j=1}^{i-1} \left( 1-\alpha_j\right) \right),
\end{equation*}
where $\alpha_i = \exp \left( - \delta_i \sigma_i \right)$ and $\delta_i$ is the distance map between planes $i$ and $i+1$.

\noindent
\textbf{Blending multiple MPI.}
To avoid frequent switching artifacts between the nearest MPI to another, we blend $k$ nearest MPI volumes ($K=3$ by default). Given a $k_\text{th}$ rendered color $\hat{\mathbf{c}}_k$ and a $k_\text{th}$ rendered alpha value $\hat{\alpha}$, we blend them on the image space according to the literature \cite{mildenhall2019local}.
\begin{equation*}
    \mathbf{c}_\text{mpi} = \frac{\sum_k w_k \hat{\alpha}_k \hat{\mathbf{c}}_k}{\sum_k w_k \hat{\alpha}_k},~
    \alpha_\text{mpi} = \min \left( \sum_k w_k \hat{\alpha}_k, 1 \right).
\end{equation*}
We output alpha values at this stage for a visualization purpose in Sect.~\ref{sec:vis_feedback}.
$w_k \propto \exp(-l / \gamma)$. $l$ is the $L^2$ distance between the camera location and the registered MPI. $\gamma$ is the maximum $L^2$ distance to the $K$ views.

\noindent
\textbf{Metric scale alignment.}
Due to the scale ambiguity of single-view imaging \cite{tucker2020single,han2022adaMPI}, layers of the direct MPI output lie within the $[0, 1]$ normalized range.
To match the scale with the real scene, we compute the scale factor $s$.
For $i_\text{th}$ MPI layer with size $w_i$ (width) by $h_i$ (height) by $z_i$ (distance from the camera), we scale the 3D rectangle as follows.
\begin{equation} \label{eq:rescaling}
    z'_i = s z_i,\quad w'_i = z'_i\frac{w_i}{f},\quad h'_i = z'_i\frac{h_i}{f},
\end{equation}
where $f$ denotes the focal length of the camera.

Since we have a metric depth $\mathbf{D}$ and can render a rendered depth image $\hat{\mathbf{D}}$ using MPI by substituting $d_i$ with $\mathbf{c}_i$ into the first equation, we calculate the scale difference in ratio, or $s$, between these depth images.
Inspired by the scale-invariant synthesis by Tucker and Snavel \cite{tucker2020single}, we calculate $s$ as follows.
\begin{equation} \label{eq:scale}
    s = \exp \left[ \frac{1}{|\mathbf{D}|}\sum_{(x,y,d)\in \mathbf{D}} \left( \ln\hat{\mathbf{D}}(x,y) -  \ln(d^{-1})\right) \right].
\end{equation}
This scaling is performed on the server. The outcome of MPI and $s$ is sent to the AR device over a network. The mobile device registers the scaled MPI into the scene using the camera parameters at the photographed location.

\subsection{Error Peaking Visualization for Next View Samples}\label{sec:vis_feedback}

\textbf{MPI rendering.}
After generating the first MPI, we switch the screen to a uniform black background and superimpose the MPI rendering. This straightforward approach intends to represent the current MPI rendering quality directly.

The resultant color values $\mathbf{c}'$ is an overlay of a rendered pixel color $\mathbf{c}_\text{mpi}$ over a black background color $\mathbf{c}_\text{blk} = (0,0,0)^\intercal$ weighted by the rendered alpha value $\alpha_\text{mpi}$.
\begin{equation}\label{eq:b_mpi}
    \mathbf{c}' = 
    \alpha_\text{mpi} \mathbf{c}_\text{mpi} + (1 - \alpha_\text{mpi})\mathbf{c}_\text{blk}.
\end{equation}
Users are expected to add another MPI when they see artifacts from the rendering, such as stack-card artifacts that reveal the layered structure when the display device changes the photographed location. However, this would be challenging for novice users without knowing the artifacts that vary depending on MPI generation methods. This is why we add error-peaking visualization.

\noindent
\textbf{Error peaking overlay.}
We calculate the difference between the video frame and MPI rendering in the screen space and highlight pixels with significant errors in red. This conveys to the users where and how much error exists. The users then add another MPI and check if the red pixels decrease.
\begin{equation}\label{eq:e_mpi}
    \mathbf{c}' =
    \begin{cases}
        \mathbf{c}_\text{err}, & \text{if } L(\mathbf{c}_\text{mpi}, \mathbf{c}_\text{vid}) > t \\
        \mathbf{c}_\text{vid}, & \text{otherwise}.
    \end{cases}
\end{equation}
Here, $\mathbf{c}_\text{err}$ denotes a pre-defined pixel color as error highlights, $\mathbf{c}_\text{vid}$ is a pixel color of a video frame, $L(\cdot)$ calculates the distance between two pixel values, and $t$ is a threshold.

We observed misaligned error highlights over the real image due to imperfect SLAM tracking. Therefore, we overlay the visualized errors over the MPI rendering. Users never see raw scenes but view them through a digital double.
\begin{equation}\label{eq:me_mpi}
    \mathbf{c}' =
    \begin{cases}
        \mathbf{c}_\text{err}, & \text{if } L(\mathbf{c}_\text{mpi}, \mathbf{c}_\text{vid}) > t \\
        \alpha_\text{mpi} \mathbf{c}_\text{mpi} + (1 - \alpha_\text{mpi})\mathbf{c}_\text{blk}, & \text{otherwise}.
    \end{cases}
\end{equation}

\section{Evaluation}

\subsection{User Study}
\textbf{Goals.}
We validate the capability of users to spontaneously sample views under our error-peaking visualization in a user study. We compare our system with LLFF \cite{mildenhall2019local}, which is based on plenoptic sampling theory, to see how different visualizations impact subjective and quantitative factors. We designed a repeated-measures within-subject study to identify the characteristics of the two systems, \texttt{Ours} and \texttt{LLFF}. We also evaluate the applicability of our system to view sampling in 3D Gaussian splatting (3DGS) \cite{kerbl20233dgs}.

\begin{figure*}[t]
    \centering
    \includegraphics[width=\textwidth]{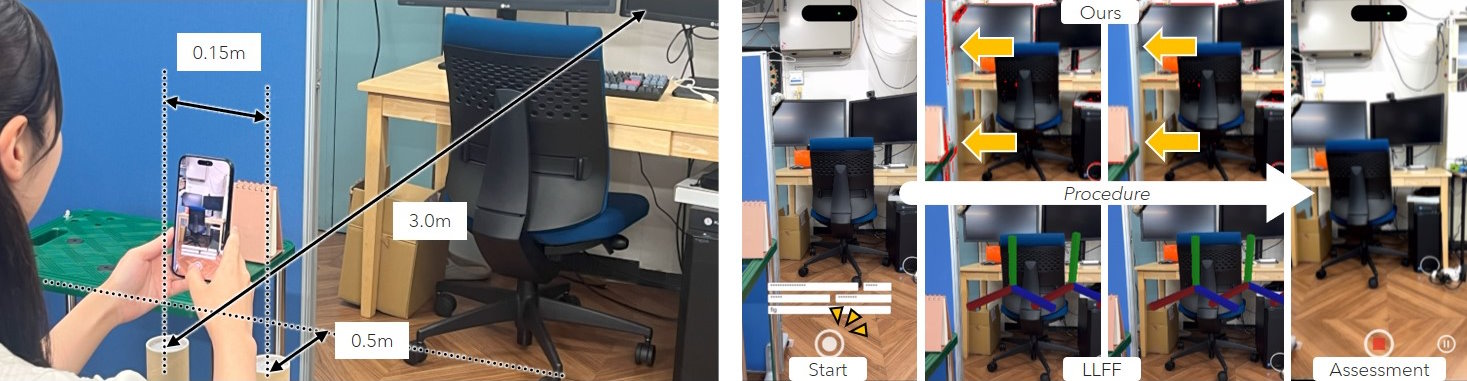}
    \caption{
    User study. 
    (Left) Experimental setup with varying depths $[0.5, 3.0]$ m.
    (Right) Experiment procedure.
    }
    \label{fig:exp_setup}
\end{figure*}

\noindent
\textbf{Scene.}
We designed an office scene with varying depths during the scene capture. The scene depth ranged approximately from 0.5 m to 3.0 m (Fig.~\ref{fig:exp_setup}). The closest depth appears when the AR device is located on the left and gradually disappears as the device moves toward the right. We placed two poles, each 0.15 m away, to regulate the capture area and inform the participants of the range.

\noindent
\textbf{Implementation details.}
We developed our mobile app for the Apple iPhone 15 Pro using Unity. We used AdaMPI but replaced the depth estimator \cite{Ranftl2021dpt} with the more recent Depth Anything v2 \cite{depth_anything_v2} for accuracy. Each MPI was $W \times H \times D = 294 \times 639 \times 32$ pixels, resulting in a $45.93\degree$ horizontal FOV in portrait mode. To avoid exceeding the limited onboard RAM, we allocated memory space only for the $k$ nearest views and swapped the content when the combination of $k$ views changed. $L(\cdot)$ calculated the L1 norm of the RGB values, and $t = 0.4$.
Sampling one MPI took $3.6$ s (SD$=0.5$).

\noindent
\textbf{Baseline implementation.}
We compare our approach with the LLFF visual guidance \cite{mildenhall2019local}, which shows 3D axes on a 2D grid based on plenoptic sampling theory. LLFF asks users to scan the entire scene in advance to calculate the minimum depth $z_{min}$ and the minimum number of views $N$ to determine camera intervals $\Delta_u$.
\begin{equation}
    N = \left( \frac{SW}{2Dz_{min}\tan\left( \theta / 2\right)}\right)^2,~\Delta_{u} = S / \sqrt{N},
\end{equation}
where $S$ is the side length of the capture area (i.e., 0.15 m in our setup) and $\theta$ is the camera field of view. Note that our approach does not require such preparation by design.

\noindent
\textbf{Metrics.}
We evaluated the systems using the System Usability Score (SUS) \cite{Brooke1996SUS}, NASA-TLX \cite{HART1988NASA}, time completion time (TCT), net completion time (NCT) excluding communication time, and the number of captured images (capture count). We also collected three additional scores from our task-oriented questionnaires: Self-Confidence (S-Conf.): \textit{“Q. The captured results will meet my expectations.”}; Satisfaction (Satis.): \textit{“Q. The captured results met my expectations.”}; and Scene Focus: \textit{“I focused on the scene of interest while capturing.”} on a 7-point scale from (1) Strongly disagree to (7) Strongly agree.

\noindent
\textbf{Participants.}
We collected 20 participants (five female and 15 male, $\Bar{X}=22.4$ (SD$=1.4$) years old, all right-handed and corrected vision). All participants were university students majoring in computer science and rated themselves in AR experience as $\Bar{X}=3.7$, SD$=1.9$ in $[1, 5]$.

\begin{figure}[t]
    \centering
    \includegraphics[width=\columnwidth]{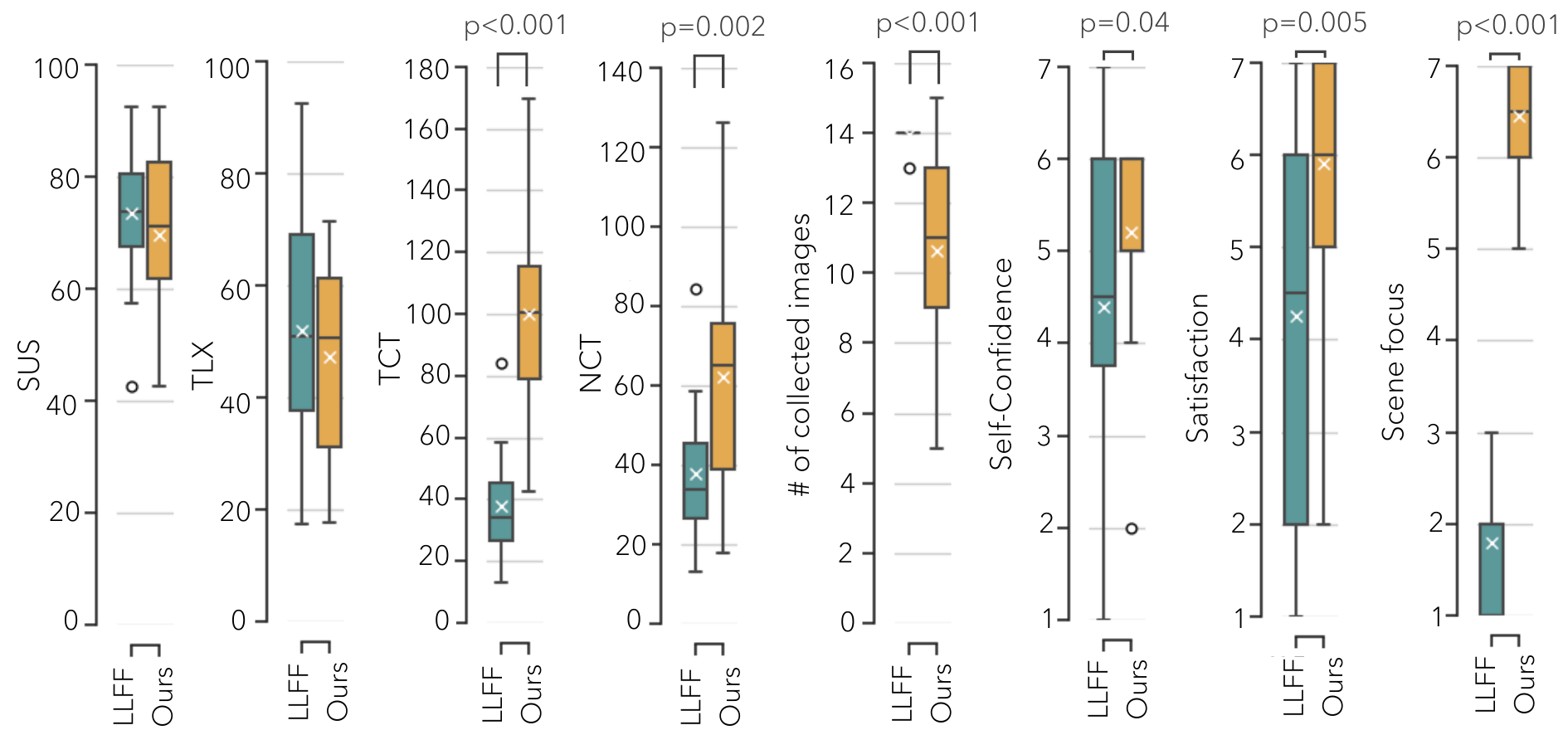}
    \caption{User study results.
    }
    \label{fig:exp_scores}
\end{figure}

\noindent
\textbf{Results.}
We performed sphericity, normality, and homogeneity of variance tests on our collected data. We used the T-test only when all the preliminary tests were satisfied, and otherwise, we used Wilcoxon signed-rank tests (Fig.~\ref{fig:exp_scores}).
The statistical analysis revealed significant differences in TCT (\texttt{LLFF}: $\Bar{X}=37.6$, SD$=15.6$; \texttt{Ours}: $\Bar{X}=99.9$, SD$=31.7$; $p<0.001$; Cohen's d$=2.4$), NCT (\texttt{LLFF}: $\Bar{X}=37.6$, SD$=15.6$; \texttt{Ours}: $\Bar{X}=62.2$, SD$=27.70$; $p=0.02$; Cohen's d$=1.1$),
the number of collected images (\texttt{LLFF}: $M=14.0$; \texttt{Ours}: $M=11.0$; $p<0.001$; RBC $=0.9$), S-Conf. (\texttt{LLFF}: $M=4.5$; \texttt{Ours}: $M=5.0$; $p=0.04$; RBC $=-0.5$), Satis. (\texttt{LLFF}: $M=4.5$; \texttt{Ours}: $M=6.0$; $p=0.005$; RBC $=-0.78$), and Scene Focus (\texttt{LLFF}: $M=2.0$; \texttt{Ours}: $M=6.5$; $p<0.001$; RBC $=-1.0$).

The analysis revealed significant differences in SUS-Q4 (LLFF: $M=2.0$; Ours: $M=3.0$; $p=0.004$; RBC $=-0.9$), SUS-Q5 (LLFF: $M=3.0$; Ours: $M=4.0$; $p=0.01$; RBC $=-0.9$), SUS-Q10 (LLFF: $M=2.0$; Ours: $M=2.0$; $p=0.03$; RBC $=-0.7$), TLX-Temporal Demand: $M=32.5$; Ours: $M=15.0$; $p=0.02$; RBC $=0.7$), and TLX-Frustration: $\Bar{X}=42.0$, SD$=28.0$; Ours: $\Bar{X}=28.5$, SD$=16.6$; $p=0.04$; Cohen's d$=0.6$).

Overall, \texttt{Ours} gives higher self-confidence during view sampling with better scene focus and lower disappointment in the final rendering quality with fewer views. However, it requires longer sessions and learning. The results validate the statements (i--iii) in Sect. \ref{sec:intro}.

\subsection{Quantitative Evaluation}
\label{sec:exp2}

\textbf{Dataset.}
We recorded a real image dataset of three scenes with varying depths and camera motions to validate statement (iv) in Sect. \ref{sec:intro}. We evaluated the rendering quality of MPI and 3DGS \cite{kerbl20233dgs} to validate the scalability of our approach. We recorded scenes densely and selected subsets to form the following virtual approaches:
(\texttt{Ours}) We calculated the mean error value $(=4.28\%)$ over all viewpoints and participants in the user study. We recorded images every time the error value exceeded this threshold.
(\texttt{Uniform}) We recorded all images and used the same number of images as in \texttt{Ours}, maintaining equal distances between views as much as possible.
(\texttt{Random}) We randomly selected the same number of images from \texttt{Uniform} and \texttt{Ours}.
We used COLMAP \cite{schoenberger2016sfm} for camera parameters and 3D points for 3DGS. Ground truth images were located between every two views of \texttt{Uniform}.

\begin{figure}[t]
    \centering
    \includegraphics[width=\columnwidth]{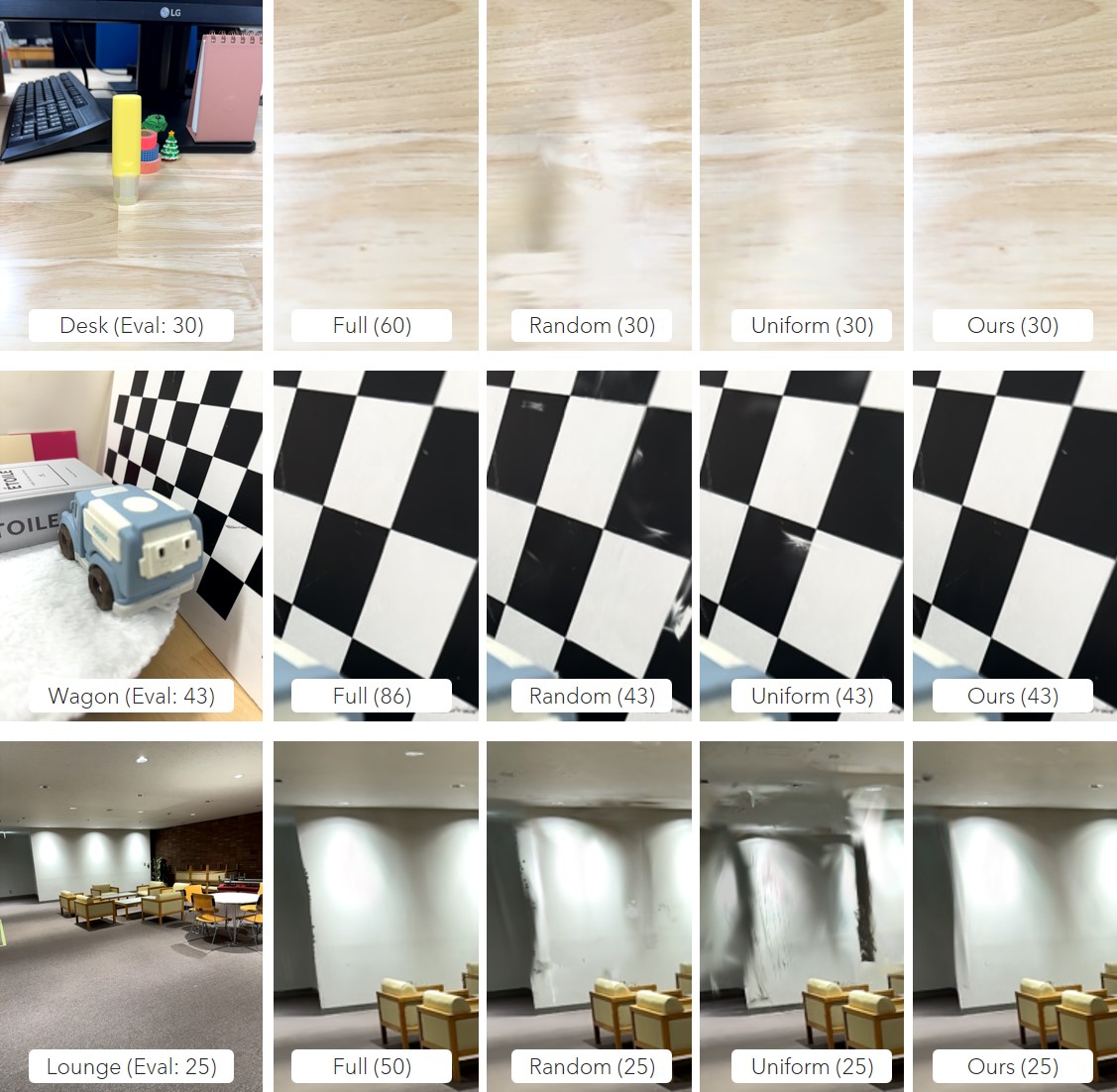}
    \caption{3DGS rendering results.}
    \label{fig:3dgs_results}
\end{figure}

\noindent
\textbf{Results.}
Given the strong dependencies on scenes, capturing strategies, and reconstruction methods, we calculated the ratios of image metrics of our competitors relative to ours, rather than using absolute image quality metrics (Table \ref{tab:results_synth}).
\texttt{Ours} outperforms the other approaches in PSNR, SSIM, and LPIPS ratios. The results in 3DGS suggest that our approach can be used as instant local feedback for other view synthesis approaches for large scenes.
Fig.~\ref{fig:3dgs_results} presents qualitative comparisons of the 3DGS results.
Here, we put ``Full'' view synthesis with all available view samples, excluding the evaluation dataset.
Our approach produces fewer artifacts than \texttt{Uniform} and \texttt{Random}, owing to error analysis by MPI, which identifies potential erroneous pixels.

\begin{table}
    \centering
    \caption{View synthesis quality of different view sampling strategies. Evaluated as ratios to account for variability in scenes, capture methods, and reconstruction techniques.}
    \label{tab:results_synth}
    \vspace{0.1em}
    \small{
    \begin{tabular}{lccc}
        \toprule
        Method (MPI) & PSNR $(\uparrow)$ & SSIM $(\uparrow)$ & LPIPS $(\downarrow)$\\
        \midrule
        Random & 0.89 (0.73) & 0.92 (0.56) & 1.09 (0.67) \\
        Uniform & 0.97 (0.87) & 0.96 (0.78) & 1.03 (0.86) \\
        Ours & 1.00 (1.00) & 1.00 (1.00) & 1.00 (1.00) \\
        \midrule
        \midrule
        Method (3DGS) & PSNR $(\uparrow)$ & SSIM $(\uparrow)$ & LPIPS $(\downarrow)$\\
        \midrule
        Random & 0.92 (1.74) & 0.986 (1.98) & 1.30 (1.96) \\
        Uniform & 0.96 (1.27) & 0.997 (1.12)  & 1.09 (1.28) \\
        Ours & 1.00 (1.00) & 1.00 (1.00) & 1.00 (1.00) \\
        \bottomrule
    \end{tabular}
    }
\end{table}

\section{Limitations and Future Directions}

Our user study and quantitative evaluation results characterize our error-peaking visualization using local MPI and light field reconstruction. However, we discuss several known limitations that the current implementation faces.

\noindent
\textbf{Depth on surfaces.}
We used Depth Anything v2 as the input depth map for AdaMPI. We found that the neural network-based depth estimator always recovers depth pixels on the surface regardless of the object materials, falling short in view synthesis for metallic and transparent objects. Error visualization can be unnecessarily significant in such areas, and users would mitigate the artifacts by taking more images.

\noindent
\textbf{Latency in communication and network inference.}
The advantage of our approach lies in real-time error visualization through view synthesis. Existing single-view MPI generation methods are designed for desktop GPUs and cannot be practically implemented on mobile capturing systems. Therefore, we attempted to implement it on a server-client system to offload the network inference while rendering from the AR device. However, our system still experiences delays due to network inference, in addition to the time required for communication between the server and the client.
Further performance improvements would lead to superior results.

\noindent
\textbf{Further user analysis.}
The sampling theory described in the literature \cite{mildenhall2019local} ignores our AR display configuration. Therefore, the theory does not explain our findings well on its own. The missing factors are the screen pixel resolution, the distance from the AR display to the users' eyes, and the users' motion while observing the rendering results. Analyzing eye-tracking data would provide more explanatory measures by quantifying the distance and the users' focus on the screen.

\section{Conclusion}

In this paper, we discussed issues in existing AR annotations for view sampling in novel view synthesis. To address these issues, we proposed a user-in-the-loop approach that tasks users with diminishing error highlights of underlying MPI reconstruction by inserting new views, rather than relying on restrictive sampling theory. Our results show that the error-peaking visualization is less invasive, reduces disappointment in final results, and is satisfactory with fewer views in our mobile view synthesis system. Additionally, our results demonstrate that photos taken with our system can benefit modern 3D Gaussian splatting.

\vspace{0.5em}
\noindent
\ifblind
\else
    \textbf{Acknowledgements} This work was supported by the Deutsche Forschungsgemeinschaft (DFG, German Research Foundation) under Germany's Excellence Strategy – EXC 2120/1 – 390831618 and partly by a grant from JST Support for Pioneering Research Initiated by the Next Generation (\# JPMJSP2123).
\fi

\bibliographystyle{IEEEbib}
\bibliography{strings}

\end{document}